\documentclass{article} 

\usepackage{amssymb}
\usepackage{graphicx}

\usepackage{subfigure} 

\usepackage{natbib}

\usepackage[ruled]{algorithm2e}

\usepackage{hyperref}

\usepackage{amsmath}
\usepackage{amsfonts}
\usepackage{enumerate}

\newcommand{\fracpartial}[2]{\frac{\partial #1}{\partial  #2}}
\newcommand{\bmat}{\begin{pmatrix}}
\newcommand{\emat}{\end{pmatrix}}

\newcommand{\R}{\mathbb{R}}

\newcommand{\Expectation}{\mathbb{E}}
\newcommand{\Var}{\mathbb{V}\operatorname{ar}}

\newcommand{\Cov}{\mathbf{\Sigma}}

\newcommand{\transp}{^{\top}}

\newcommand{\params}{\boldsymbol{\theta}}
\newcommand{\grad}{\nabla_{\params}}

\newcommand{\gradj}{\nabla_{\theta_{i}}}

\newcommand{\hess}{\mathbf{H}}
\newcommand{\cent}{\mathbf{c}}

\newcommand{\hatg}{\overline{g_i}}
\newcommand{\hatv}{\overline{v_i}}
\newcommand{\hath}{\overline{h_i}}
\newcommand{\hatvg}{\overline{l}}

\newcommand{\adagrad}{{\sc AdaGrad}}

\usepackage{color}
\definecolor{dkgreen}{rgb}{0.1,0.4,0}
\definecolor{orange}{rgb}{0.8,0.4,0}
\definecolor{nyupurple}{rgb}{0.5,0.0,0.9}
\definecolor{violet}{rgb}{0.9,0.,0.5}

\usepackage[accepted]{icml2013}

\icmltitlerunning{No More Pesky Learning Rates}

\begin{document}

\twocolumn[
\icmltitle{No More Pesky Learning Rates}

\icmlauthor{Tom Schaul}{schaul@cims.nyu.edu}
\icmlauthor{Sixin Zhang}{zsx@cims.nyu.edu}
\icmlauthor{Yann LeCun}{yann@cims.nyu.edu}
\icmladdress{Courant Institute of Mathematical Sciences\\
New York University\\
715 Broadway, New York, NY 10003, USA}

\icmlkeywords{stochastic gradient descent, adaptation, learning rates, neural networks, non-stationary, online learning}

\vskip 0.3in
]

\begin{abstract}
The performance of stochastic gradient descent (SGD) depends
critically on how learning rates are tuned and decreased over time.
We propose a method to automatically adjust multiple learning rates so
as to minimize the expected error at any one time. The method relies
on local gradient variations across samples. 
In our approach, learning rates can increase as well as decrease, 
making it suitable for non-stationary problems.
Using a number of convex
and non-convex learning tasks, we show that the resulting algorithm
matches the performance of SGD or other adaptive approaches with their best settings obtained through
systematic search, and effectively removes the need for learning rate
tuning. 
\end{abstract}

\section{Introduction}
Large-scale learning problems require algorithms that scale benignly
(e.g. sub-linearly) with the size of the dataset and the number of
trainable parameters. This has lead to a recent resurgence of interest
in \emph{stochastic gradient descent} methods (SGD). Besides fast
convergence, SGD has sometimes been observed to yield significantly
better generalization errors than batch methods~\cite{bottou-bousquet-11}.

In practice, getting good performance with SGD requires some manual
adjustment of the initial value of the learning rate (or step size)
for each model and each problem, as well as the design of an annealing
schedule for stationary data. The problem is particularly acute for
non-stationary data.

The contribution of this paper is a novel method to
\emph{automatically adjust} learning rates (possibly different
learning rates for different parameters), so as to minimize some
estimate of the expectation of the loss at any one time. 

Starting from an idealized scenario where every sample's contribution
to the loss is quadratic and separable, we derive a formula for the
optimal learning rates for SGD, based on estimates of the variance of
the gradient.  The formula has two components: one that captures
variability across samples, and one that captures the local
curvature, both of which can be estimated in practice. The method can
be used to derive a single common learning rate, or local learning
rates for each parameter, or each block of parameters, leading to five
variations of the basic algorithm, none of which need any parameter
tuning.

The performance of the methods {\em obtained without any manual
  tuning} are reported on a variety of convex and non-convex learning
models and tasks. They compare favorably with an ``ideal SGD'', where
the best possible learning rate was obtained through systematic
search, as well as previous adaptive schemes.


\section{Background}
\label{sec:background}
SGD methods have a long history in adaptive signal processing, neural
networks, and machine learning, with an extensive literature
(see~\cite{bottou-98x,bottou-bousquet-11} for recent reviews). While
the practical advantages of SGD for machine learning applications have
been known for a long time~\cite{lecun-98b}, interest in SGD has
increased in recent years due to the ever-increasing amounts of
streaming data, to theoretical optimality results for generalization
error~\cite{bottou-lecun-04}, and to competitions being won by SGD
methods, such as the PASCAL Large Scale Learning
Challenge~\cite{bordes-jmlr-09}, where Quasi-Newton approximation
of the Hessian was used within SGD.  Still, practitioners need to deal
with a sensitive hyper-parameter tuning phase to get top performance:
each of the PASCAL tasks used very different parameter settings.  This
tuning is very costly, as every parameter
setting is typically tested over multiple epochs.

Learning rates in SGD are generally decreased according a schedule of
the form $\eta(t) = \eta_0 (1+\gamma t)^{-1}$. Originally proposed as
$\eta(t) = O(t^{-1} )$ in~\cite{robbins-monro-51}, this form was
recently analyzed in~\cite{xu-10,bach-nips-11} from a non-asymptotic
perspective to understand how hyper-parameters like $\eta_0$ and
$\gamma$ affect the convergence speed.  

Numerous researchers have proposed schemes for making learning rates
adaptive, either globally or by adapting one rate per parameter (`diagonal preconditioning');
see~\cite{George2006} for an overview. 
An early diagonal preconditioning schemes was proposed in \cite{Almeida1999}
where the learning rate is adapted as 
\begin{equation*}
	\eta_{i} (t) = \max\left(0,\frac {\eta_0 \; \theta_i(t) \cdot \nabla_{\theta_i}^{(t-1)} } { \hatv} \right)
\end{equation*}
for each problem dimension $i$, where 
$\nabla_{\theta_i}^{(t)}$ is gradient of the $i$th parameter at iteration $t$, and $\hatv \approx \Expectation\left[  \nabla_{\theta_i}^{2} \right]$ is a recent running average of its square.
Stochastic meta-descent (SMD,~\citet{schraudolph1999local,schraudolph2002fast}) 
uses a related multiplicative update of learning rates.
Approaches based on the natural gradient \cite{Amari2000} precondition the updates by the 
empirical Fisher information matrix (estimated by the gradient covariance matrix, or its diagonal approximation), in the simplest case: $\eta_i = \eta_0 / \hatv$; the ``Natural Newton'' algorithm~\cite{le2010fast} combines the gradient covariance with second-order information.
Finally, derived from a worst-case analysis, \cite{DuchiHS11} propose an approach called `\adagrad', where the learning rate takes the form 
\begin{equation*}
	\eta_{i} (t) =\frac { \eta_0 } { \sqrt{ \sum_{s=0}^{t} { \left(\nabla_{\theta_i}^{(s)} \right)}^2 }}.
\end{equation*}
The main practical drawback for all of these approaches is that they retain one or more sensitive hyper-parameters,
which must be tuned to obtain satisfactory performance.
\adagrad\ has another disadvantage: because it accumulates all the gradients from the moment training starts to determine the current learning rate, the learning rate monotonically decreases: this is especially problematic for non-stationary problems, but also on stationary ones,
as navigating the properties of optimization landscape change continuously.

The main contribution of the present paper is {\em a formula that
  gives the value of the learning rate that will maximally decrease
  the expected loss after the next update}. For efficiency reasons,
some terms in the formula must be approximated using such quantities
as the mean and variance of the gradient. As a result, the
  learning rate is {\em automatically decreased to zero} when approaching an
  optimum of the loss, without requiring a pre-determined annealing
  schedule, and if the problem is non-stationary, it the learning rate \emph{grows again} when the data changes.


\section{Optimal Adaptive Learning Rates}
In this section, we derive an optimal learning
rate schedule, using an idealized quadratic and separable loss
function. We show that using this learning rate schedule preserves
convergence guarantees of SGD.  In the following section, we find how the optimal learning
rate values can be estimated from available information, and describe
a couple of possible approximations.

The samples, indexed by $j$, are drawn i.i.d. from a data distribution
$\mathcal{P}$. Each sample contributes a {\em per-sample loss}
$\mathcal{L}^{(j)}(\params)$ to the expected loss:
\begin{eqnarray}
	\mathcal{J}(\params)  = \Expectation_{j \sim \mathcal{P}} \left[ \mathcal{L}^{(j)}(\params) \right]
\end{eqnarray}
where $\params \in \R^d$ is the trainable parameter vector, whose optimal value 
is denoted $\params^{*} = \arg\min_{\params}\mathcal{J}(\params)$.
The SGD parameter update formula is of the form  
$\params^{(t+1)} = \params^{(t)} - \eta^{(t)} \grad^{(j)}$,
where $\grad^{(j)} =  \fracpartial{}{\params}\mathcal{L}^{(j)}(\params) $ 
is the gradient of the the contribution of example $j$ to the loss,
and the learning rate $\eta^{(t)}$ is a suitably chosen sequence of
positive scalars (or positive definite matrices).

\subsection{Noisy Quadratic Loss}

\begin{figure}[tb]
\centerline{
\includegraphics[width=\columnwidth]{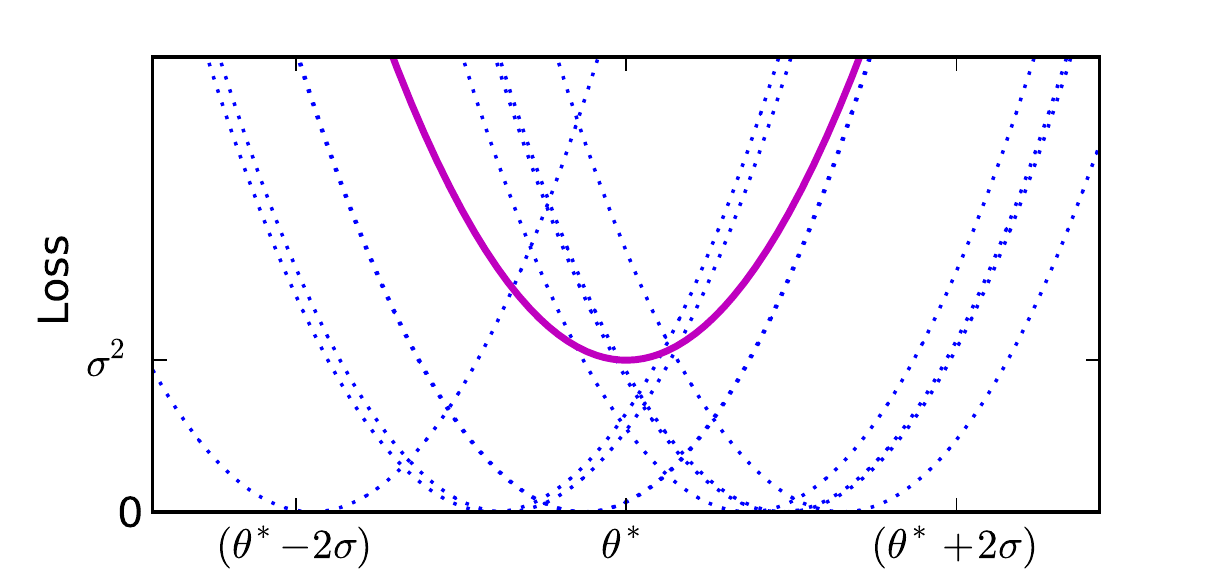}
}
\caption{Illustration of the idealized loss function considered (thick
  magenta), which is the average of the quadratic contributions of
  each sample (dotted blue), with minima distributed around the point
  $\theta^*$. Note that the curvatures are assumed to be identical
  for all samples.
}
\label{fig:illust}%
\end{figure}

We assume that the per-sample loss functions are smooth around minima,
and can be locally approximated by a quadratic function. We also
assume that the minimum value of the per-sample loss functions are zero:
\begin{eqnarray*}
\mathcal{L}^{(j)}(\params) = \frac{1}{2}\left(\params-\cent^{(j)}\right)\transp\hess^{(j)}\left(\params-\cent^{(j)}\right) 
 \\
 \grad^{(j)} = \hess^{(j)}\left(\params-\cent^{(j)}\right)
\end{eqnarray*}
where $\hess_i$ is the (positive semi-definite) Hessian matrix
of the per-sample loss of sample $j$, and $\cent^{(j)}$ is the
optimum for that sample.  The distribution of per-sample optima
$\cent^{(j)}$ has mean $\params^*$ and variance $\Cov$.
Figure~\ref{fig:illust} illustrates the scenario in one dimension.

To simplify the analysis, we assume for the remainder of this section
that the Hessians of the per-sample losses are identical for all
samples, and that the problem is separable, i.e., the Hessians are
diagonal, with diagonal terms denoted $\{h_1,\ldots,h_i,\ldots,h_d\}$.
Further, we will ignore the off-diagonal terms of $\Cov$, and
denote the diagonal
$\{\sigma^2_1,\ldots,\sigma^2_i,\ldots,\sigma^2_d\}$.  Then, for any
of the $d$ dimensions, we thus obtain a one-dimensional problem (all
indices $i$ omitted).
\begin{eqnarray}
J(\theta) = \Expectation_{i \sim \mathcal{P}}\left[\frac{1}{2}h(\theta-c^{(j)})^2\right] 
 =  \frac{1}{2}h  \left[\left(\theta- \theta^*\right)^2 + \sigma^2 \right]
 \label{eq:1dloss}
\end{eqnarray}
The gradient components are $\nabla_{\theta}^{(j)} = h\left(\theta - c^{(j)}\right)$,
with
\begin{eqnarray}
\Expectation [\nabla_{\theta}] = h (\theta-\theta^*)
\label{eq:grad-exp}
&&
\Var[\nabla_{\theta}] = h^2\sigma^2
\end{eqnarray}
and we can rewrite the SGD update equation as
%
%
\begin{eqnarray}
\theta^{(t+1)} 
&=& \theta^{(t)} - 
\eta  h
\left(\theta^{(t)} - c^{(j)}\right)
\nonumber \\
&=&
(1-\eta h  )\theta^{(t)} + \eta h\theta^*
+ \eta h \sigma    \xi^{(j)}
\label{eq:update}
\end{eqnarray}
where the $\xi^{(j)}$ are i.i.d. samples from a zero-mean and unit-variance Gaussian distribution.
%
Inserting this into equation~\ref{eq:1dloss}, we obtain the expected
loss after an SGD update
\begin{eqnarray*}
&&\Expectation \left[J\left(\theta^{(t+1)}\right) \; | \;\theta^{(t)}\right] \\
&=& \frac{1}{2}h \cdot
\left[
(1-\eta h  )^2 (\theta^{(t)} -\theta^*)^2  + \eta^2 h^2 \sigma^2 +\sigma^2
\right]
\end{eqnarray*}

\subsection{Optimal Adaptive Learning Rate}
We can now derive the optimal (greedy) learning rates for the current
time $t$ as the value $\eta^* (t)$ that minimizes the expected loss
after the next update
\begin{eqnarray}
\eta^* (t) \hspace{-0.5cm}&&= \arg\min_{\eta} \left[ (1-\eta h  )^2 (\theta^{(t)} -\theta^*)^2 
+\sigma^2 + \eta^2 h^2 \sigma^2 \right] \nonumber\\
 && =\arg\min_{\eta} \left[\eta^2 \left( h (\theta^{(t)} -\theta^*)^2 +  h \sigma^2\right) 
\right. \nonumber\\ && \hspace{2cm} \left.
-2\eta   (\theta^{(t)} -\theta^*)^2 \right] \nonumber \\
&&= \frac{1}{h} \cdot \frac{(\theta^{(t)} -\theta^*)^2} { (\theta^{(t)} -\theta^*)^2 + \sigma^2}
\label{eq:opt-lr}
\end{eqnarray}
In the classical (noiseless or batch) derivation of the optimal
learning rate, the best value is simply $ \eta^* (t) = h^{-1}$. The
above formula inserts a corrective term that {\em reduces} the
learning rate whenever the sample pulls the parameter vector in
different directions, as measured by the gradient variance
$\sigma^2$. The reduction of the learning rate is larger near an
optimum, when $(\theta^{(t)} -\theta^*)^2$ is small relative to
$\sigma^2$. In effect, this will reduce the expected error due to the
noise in the gradient.  Overall, this will have the same effect as the
usual method of progressively decreasing the learning rate as we get
closer to the optimum, {\em but it makes this annealing schedule
  automatic}.


If we do gradient descent with $\eta^*(t)$, then almost surely, the algorithm
converges (for the quadratic model). The proof is given in the appendix.

\subsection{Global vs. Parameter-specific Rates}
The previous subsections looked at the optimal learning rate in the
one-dimensional case, which can be trivially generalized to $d$
dimensions if we assume that all parameters are separable, namely by
using an individual learning rate $\eta^*_i$ for each dimension $i$.
Alternatively, we can derive an optimal \emph{global} learning rate
$\eta^*_g$ (see appendix for the full derivation), 
\begin{eqnarray}
\eta_g^*(t)
&=& 
\frac{\sum_{i=1}^d h^2_i  (\theta_i^{(t)} -\theta^*_i)^2}
{ \sum_{i=1}^d \left(
h_i^3 (\theta_i^{(t)} -\theta^*_i)^2 
+  h_i^3 \sigma_i^2\right)
}
\label{eq:opt-lr-g}
\end{eqnarray}
which is especially useful if the problem is badly conditioned.

In-between a global and a component-wise learning rate, it is of
course possible to have common learning rates for blocks of
parameters. In the case of multi-layer learning systems, the blocks
may regroup the parameters of each single layer, the biases, etc.
This is particularly useful in \emph{deep} learning, where the gradient magnitudes can vary
significantly between shallow and deep layers.

\section{Approximations}
In practice, we are not given the quantities $\sigma_i$, $h_i$ and
$(\theta_i^{(t)} -\theta_i^*)^2$.  However, based on
equation~\ref{eq:grad-exp}, we can estimate them from the observed
samples of the gradient:
\begin{equation}
\eta_i^* = \frac{1}{h_i} \cdot \frac{\left(\Expectation [\nabla_{\theta_i}]\right)^2}
{\left(\Expectation [\nabla_{\theta_i}]\right)^2 + \Var[\nabla_{\theta_i}] }
= \frac{1}{h_i} \cdot \frac{\left(\Expectation [\nabla_{\theta_i}]\right)^2}
{\Expectation [\nabla_{\theta_i}^2]}
\label{eq:opt-approx}
\end{equation}

The situation is slightly different for the global learning rate
$\eta^*_g$. Here we assume that it is feasible to estimate the maximal
curvature $h^+ = \max_i(h_i)$ (which can be done efficiently, for
example using the diagonal Hessian computation method described in
\cite{lecun-98b}).  Then we have the bound
\begin{eqnarray}
\eta_g^*(t) & \geq &
\frac{1}{h^+} \cdot
\frac{\sum_{i=1}^d h^2_i  (\theta_i^{(t)} -\mu_i)^2}
{ \sum_{i=1}^d \left(
h_i^2 (\theta_i^{(t)} -\mu_i)^2 
+ h_i^2 \sigma_i^2\right)
}
\nonumber \\
&=&
\frac{1}{h^+} \cdot
\frac{\left\|\Expectation [\nabla_{\params}] \right\|^2}
{\Expectation \left[ \left\| \nabla_{\params}\right\|^2\right] 
}
\label{eq:opt-approx-g}
\end{eqnarray}
because
\begin{equation*}
\Expectation \left[ \left\| \nabla_{\params}\right\|^2\right] = \Expectation \left[ \sum_{i=1}^d (\nabla_{\theta_i})^2\right] =   \sum_{i=1}^d \Expectation\left[(\nabla_{\theta_i})^2\right]
\end{equation*}
In both cases (equations~\ref{eq:opt-approx}
and~\ref{eq:opt-approx-g}), the optimal learning rate is decomposed
into two factors, one term which is the inverse curvature (as is the case
for batch second-order methods), and one novel term that depends on
the noise in the gradient, relative to the expected squared norm of the
gradient.  Below, we approximate these terms separately.  For the
investigations below, when we use the true values instead of a practical
algorithm, we speak of the `oracle' variant (e.g. in Figure~\ref{fig:quad}).

\subsection{Approximate Variability}
We use an exponential moving average with time-constant $\tau$ 
(the approximate number of samples considered from recent memory)
for online estimates of the quantities in equations~\ref{eq:opt-approx}
and~\ref{eq:opt-approx-g}:
\begin{eqnarray*}
\hatg (t+1) &=& (1-\tau_i^{-1}) \cdot \hatg(t) + \tau_i^{-1} \cdot \nabla_{\theta_i(t)}\\
\hatv (t+1) &=& (1-\tau_i^{-1}) \cdot \hatv(t) + \tau_i^{-1} \cdot (\nabla_{\theta_i(t)})^2\\
\hatvg (t+1) &=& (1-\tau^{-1}) \cdot \hatvg(t) + \tau^{-1} \cdot \left\| \nabla_{\params}\right\|^2
\end{eqnarray*}
where $\hatg$ estimates the average gradient component $i$,
$\hatv$ estimates the uncentered variance on gradient component $i$,
and $\hatvg$ estimates the squared length of the gradient vector:
\begin{eqnarray*}
\hatg  \approx \Expectation [\nabla_{\theta_i}] \hspace{1cm}
\hatv  \approx \Expectation [\nabla_{\theta_i}^2]\hspace{1cm}
\hatvg \approx \Expectation \left[ \left\| \nabla_{\params}\right\|^2\right]
\end{eqnarray*}
and we need $\hatv$ only for an element-wise adaptive learning rate
and $\hatvg$ only in the global case.

\begin{algorithm}[tb]
\DontPrintSemicolon
\SetKwInOut{Input}{input}
\SetKwInOut{Output}{output}
\label{alg:vdSGD}
\caption{Stochastic gradient descent with adaptive learning rates (element-wise, vSGD-l).}
 \Repeat{stopping criterion is met}{
  draw a sample $c^{(j)}$,
  compute the gradient $\nabla_{\params}^{(j)}$, and
  compute the diagonal Hessian estimates $h_i^{(j)}$ using the ``bbprop'' method\\
  \For{$i \in \{1, \ldots,d\}$}{
  update moving averages\\
 $ \begin{array}{lll}
\hatg  &\leftarrow &(1-\tau_i^{-1}) \cdot \hatg + \tau_i^{-1} \cdot \gradj^{(j)}\\
\hatv  &\leftarrow &(1-\tau_i^{-1}) \cdot \hatv + \tau_i^{-1} \cdot \left(\gradj^{(j)}\right)^2\\
\hath  &\leftarrow &(1-\tau_i^{-1}) \cdot \hath + \tau_i^{-1} \cdot \left|\text{bbprop}(\params)^{(j)}_i \right|\\
\end{array}$\\
  estimate learning rate 
  $\;\;\displaystyle\eta_i^* \leftarrow \frac{(\hatg)^2}{\hath \cdot\hatv}$\\
 update memory size\\ $\;\;\tau_i \leftarrow \left(1-\frac{(\hatg)^2}{\hatv}\right) \cdot  \tau_i+ 1 $\\
 update parameter $\;\;\displaystyle \theta_i \leftarrow \theta_i - \eta_i^* \gradj^{(j)}$\\
 }
 }
\end{algorithm}

\subsection{Adaptive Time-constant}
We want the size of the memory to increase when the steps taken are small (increment by 1), 
and to decay quickly if a large step (close to the Newton step) is taken,
which is obtained naturally, by the following update
\begin{eqnarray*}
\tau_i (t+1) &=& \left(1-\frac{\hatg(t)^2}{\hatv(t)}\right) \cdot  \tau_i(t) \;+\; 1,
\label{eq:memory-update}
\end{eqnarray*}
This way of making the memory size adaptive allows us to eliminate one otherwise tuning-sensitive hyperparameter.
Note that these updates (correctly) do not depend on the local curvature, making them scale-invariant.

\subsection{Approximate Curvature}
There exist a number of methods for obtaining an online estimates of
the diagonal Hessian~\cite{Martens2012,bordes-jmlr-09,chapelle2011improved}.  We adopt the ``bbprop'' method, which computes
positive estimates of the diagonal Hessian terms (Gauss-Newton approximation) for a single sample
$h_i^{(j)}$, using a back-propagation formula~\cite{lecun-98b}. The
diagonal estimates are used in an exponential moving average procedure
\begin{eqnarray*}
\hath (t+1) &=& (1-\tau_i^{-1}) \cdot \hath(t) + \tau_i^{-1} \cdot h_i^{(t)}
\end{eqnarray*}
If the curvature is close to zero for some component, this can drive
$\eta^*$ to infinity.  Thus, to avoid numerical instability (to bound
the condition number of the approximated Hessian), it is possible to enforce a lower
bound $\hath \geq \epsilon$.  This addition is not necessary in our experiments,
due to the presence of an L2-regularization term.

\subsection{Slow-start Initialization}
To initialize these estimates, we compute the arithmetic averages over a handful ($n_0 =0.001 \times \mbox{\#traindata} $) of 
samples before starting to the main algorithm loop. We find that the algorithm works best with a \emph{slow start} heuristic,
where the parameter updates are kept small until the exponential averages become sufficiently accurate.
This is achieved by overestimating $\hatv$ and $\hatvg$) by a factor $C$.
We find that setting $C=d/10$, as a rule of thumb is both robust and near-optimal, because 
the value of $C$ has only a transient initialization effect on the algorithm.
The appendix details how we arrived at this, and demonstrates the low sensitivity empirically.

\section{Adaptive Learning Rate SGD}
\label{sec:vsgd}
The simplest version of the method views each component in
isolation. This form of the algorithm will be called ``vSGD'' (for
``variance-based SGD'').  In realistic settings with high-dimensional
parameter vector, it is not clear a priori whether it is best to have
a single, global learning rate (that can be estimated robustly), a
set of local, dimension-specific rates, or block-specific learning rates
(whose estimation will be less robust). We propose three variants on this spectrum:

\begin{figure}[tb]
	\centerline{
	\includegraphics[width=0.98\linewidth, clip=true, trim=0.22cm 0.15cm 0.275cm 0.15cm]{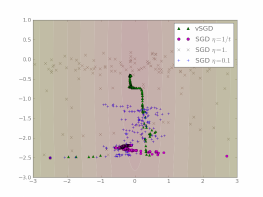}
}
	\caption{Illustration of the dynamics in a noisy
          quadratic bowl (with 10 times larger curvature in one dimension than the other).  
          Trajectories of 400 steps from vSGD, and
          from SGD with three different learning rate schedules.  SGD
          with fixed learning rate (crosses) descends until a certain
          depth (that depends on $\eta$) and then oscillates. SGD with
          a $1/t$ cooling schedule (pink circles) converges
          prematurely.  On the other hand, vSGD (green
          triangles) is much less disrupted
          by the noise and continually approaches the optimum. 
          }
	\label{fig:bowl}
\end{figure}

\begin{figure}[tb]
	\centerline{
	\vspace{-0.5em}
	\includegraphics[width=0.95\linewidth, clip=true, trim=0.5cm 1cm 1cm 1.5cm]{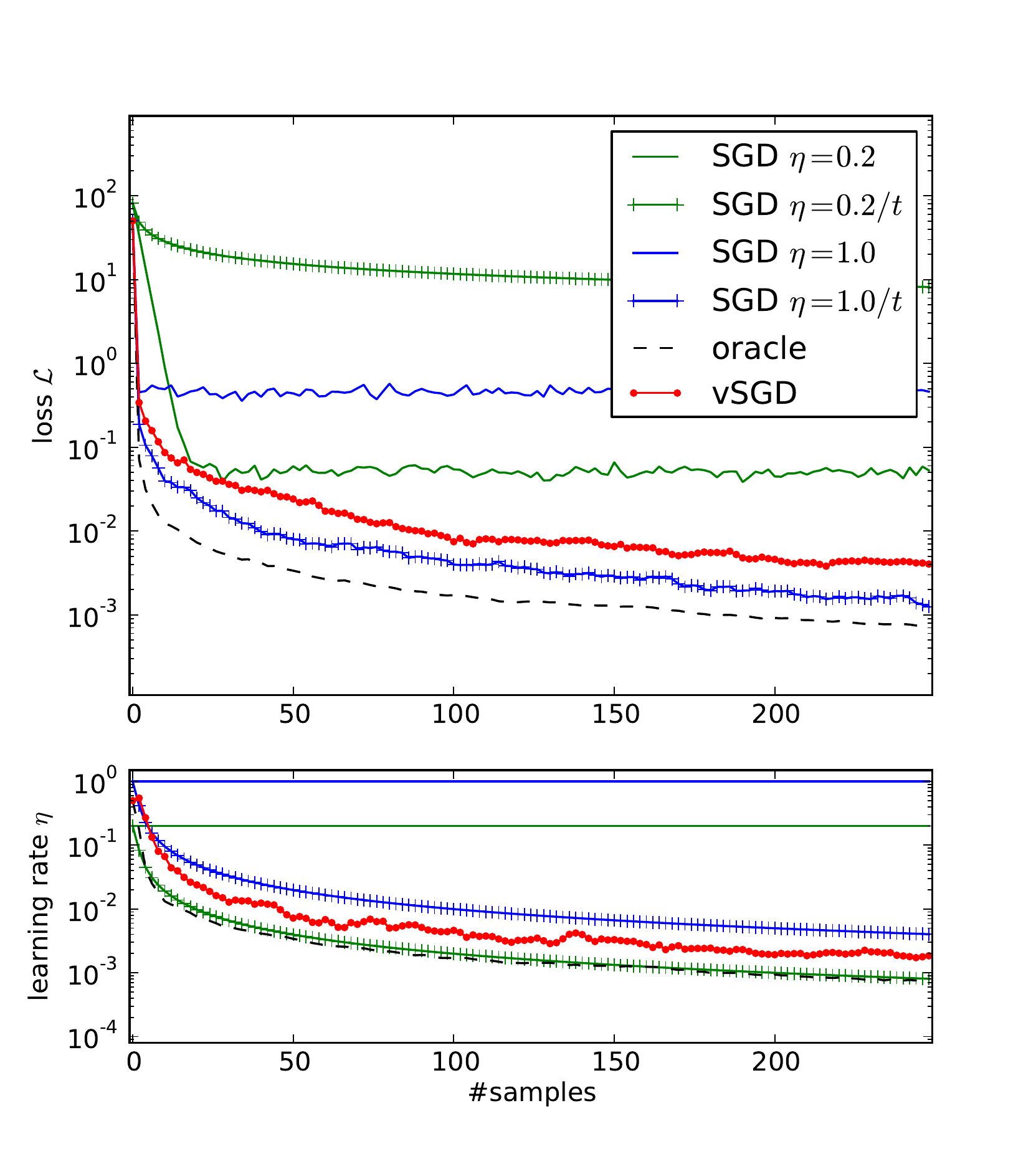}
}
	\caption{Optimizing a noisy quadratic loss
          (dimension $d=1$, curvature $h=1$). Comparison between SGD
          for two different fixed learning rates 1.0 and 0.2, and two
          cooling schedules $\eta=1/t$ and $\eta=0.2/t$, and vSGD (red
          circles).  In dashed black, the `oracle'
          computes the true optimal learning rate
          rather than approximating it.  In the top subplot, we show the median
          loss from 1000 simulated runs, and below are corresponding
          learning rates.  We observe that vSGD initially descends as
          fast as the SGD with the largest fixed learning rate, but
          then quickly reduces the learning rate which dampens the
          oscillations and permits a continual reduction in loss,
          beyond what any fixed learning rate could achieve.  The best
          cooling schedule ($\eta=1/t$) outperforms vSGD, but when the
          schedule is not well tuned ($\eta=0.2/t$), the effect on the
          loss is catastrophic, even though the produced learning
          rates are very close to the oracle's (see the overlapping
          green crosses and the dashed black line at the bottom).  }
	\label{fig:quad}
\end{figure}

\begin{description}
\item[vSGD-l] uses \emph{local} gradient variance terms
  and the local diagonal Hessian estimates,
  leading to $\eta_i^*=(\hatg)^2/(\hath \cdot\hatv)$,
\item[vSGD-g] uses a \emph{global} gradient variance term and an upper bound
  on diagonal Hessian terms: $\eta^*=\sum(\hatg)^2/(h^+ \cdot
  \hatvg)$,
\item[vSGD-b] operates like vSGD-g, but being only global across
  multiple (architecture-specific) \emph{blocks} of parameters, with a
  different learning rate per block. Similar ideas are adopted in
  TONGA~\cite{leroux-nips-08}. In the experiments, the parameters
  connecting every two layers of the network are regard as a block,
  with the corresponding bias parameters in separate blocks.
\end{description}
The pseudocode for vSGD-l is given in Algorithm~\ref{alg:vdSGD}, the other cases are very similar; all of them have linear complexity in time and space; in fact, the overhead of vSGD is roughly a factor two, which arises from the additional bbrop pass (which could be skipped in all but a fraction of the updates) -- this cost is even less critical because it can be trivially parallelized.

\begin{figure*}[tbh]
 \centering
 \includegraphics[width=0.99\textwidth, clip=true, trim=2cm 0.85cm 2cm 1.6cm]{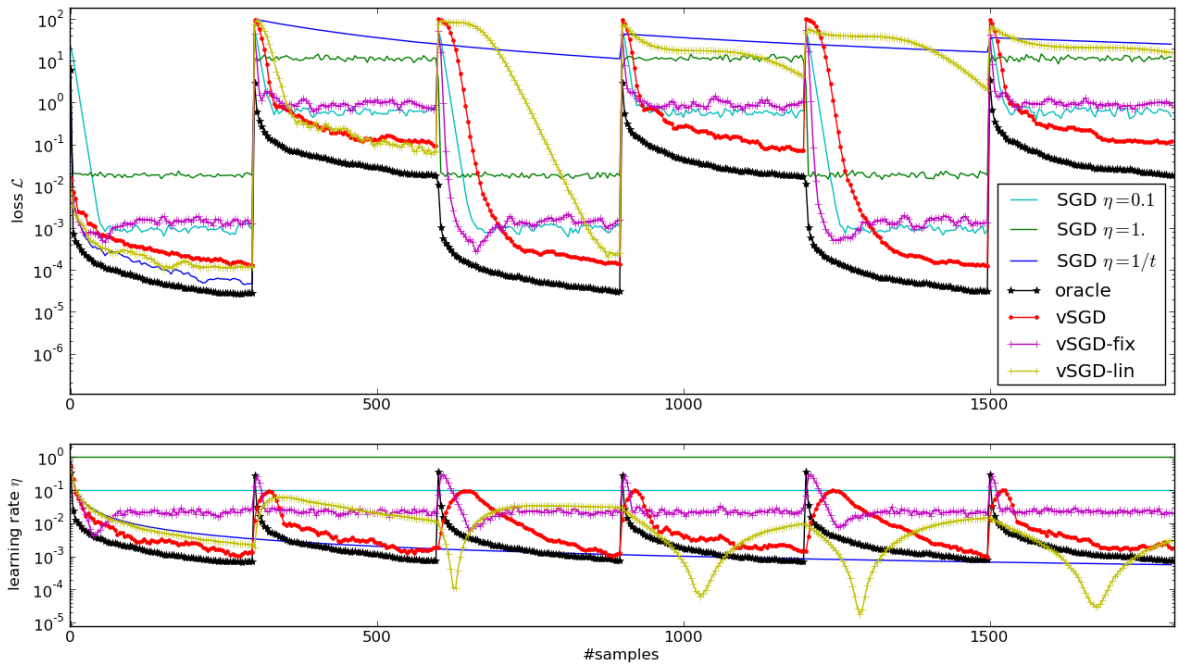}
 \caption{Non-stationary loss. The loss is quadratic but
   now the target value ($\mu$) changes abruptly every 300 time-steps.
   Above: loss as a function of time, below: corresponding learning rates.
   This illustrates the limitations of SGD with fixed or decaying
   learning rates (full lines): any fixed learning rate limits the
   precision to which the optimum can be approximated (progress
   stalls); any cooling schedule on the other hand cannot cope with
   the non-stationarity.  In contrast, our adaptive
   setting (`vSGD', red circles), as closely resembles the optimal behavior
   (oracle, black dashes). The learning rate decays like $1/t$ during the static part, but increases 
   again after each abrupt change (with just a very small delay compared to the oracle). The 
   average loss across time is substantially better than for any SGD cooling schedule.
   }
 \label{fig:nonstation}
\end{figure*}

\section{Experiments}
We test the new algorithm extensively on a couple of toy problem first,
and then follow up with results on well-known benchmark problems
for digit recognition, image classification and image reconstruction,
using the new SGD variants to train both convex models (logistic regression)
and non-convex ones (multi-layer perceptrons).

\subsection{Noisy Quadratic}
\label{sec:quad}
To form an intuitive understanding of the effects of the optimal
adaptive learning rate method, and the effect of the approximation, we
illustrate the oscillatory behavior of SGD, and compare the decrease
in the loss function and the accompanying change in learning rates on
the noisy quadratic loss function from Section 3.1 (see Figure~\ref{fig:bowl} and Figure~\ref{fig:quad}), 
contrasting the effect of fixed rates or fixed schedules to adaptive learning rates, whether in approximation or using the oracle.

\begin{figure*}[htp]
\centering
\includegraphics[width=0.96\linewidth]{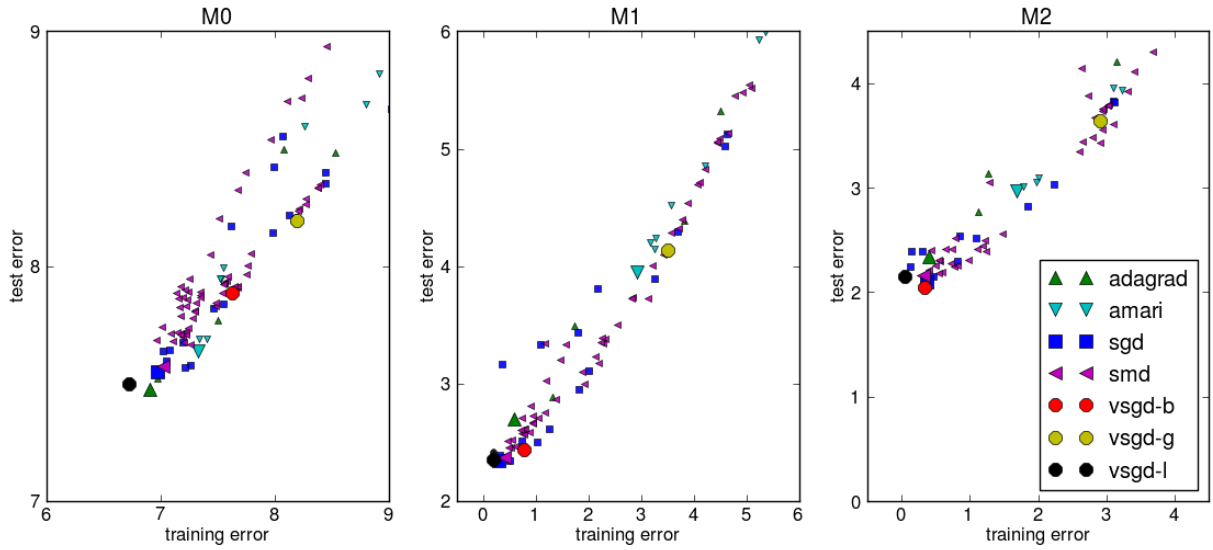}
\caption{Training error versus test error on the three MNIST setups (after 6 epochs). Different symbol-color combinations correspond to different algorithms, with the best-tuned parameter setting shown as a much larger symbol than the other settings tried (the performance of Almeida is so bad it's off the charts).
The axes are zoomed to the regions of interest for clarity, for a more global perspective, and for the corresponding plots on the CIFAR benchmarks, see Figures~\ref{fig:tvt-cifar} and~\ref{fig:tvt-glob}.
Note that there was no tuning for our parameter-free vSGD, yet its performance is consistently good (see black circles).
}
	\label{fig:tvt}
\end{figure*}

\begin{figure*}[htp]
\centering
\includegraphics[width=\linewidth]{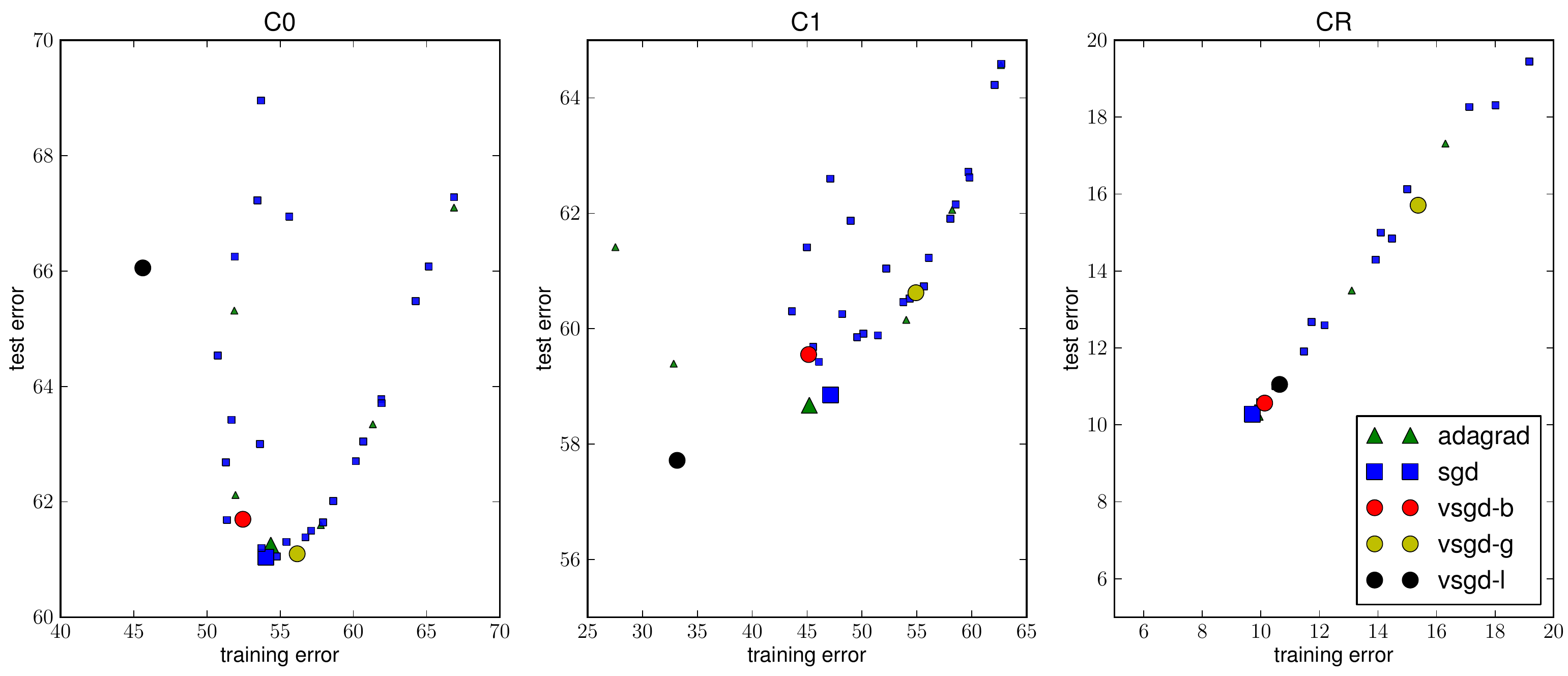}
\caption{Training error versus test error on the three CIFAR setups (after 6 epochs). Different symbol-color combinations correspond to different algorithms, with the best-tuned parameter setting shown as a much larger symbol than the other settings tried.
The axes are zoomed to the regions of interest for clarity.
Note how there is much more overfitting here than for MNIST, even with vanilla SGD.
}
	\label{fig:tvt-cifar}
\end{figure*}

\begin{figure*}[htp]
\centering
\includegraphics[width=\linewidth]{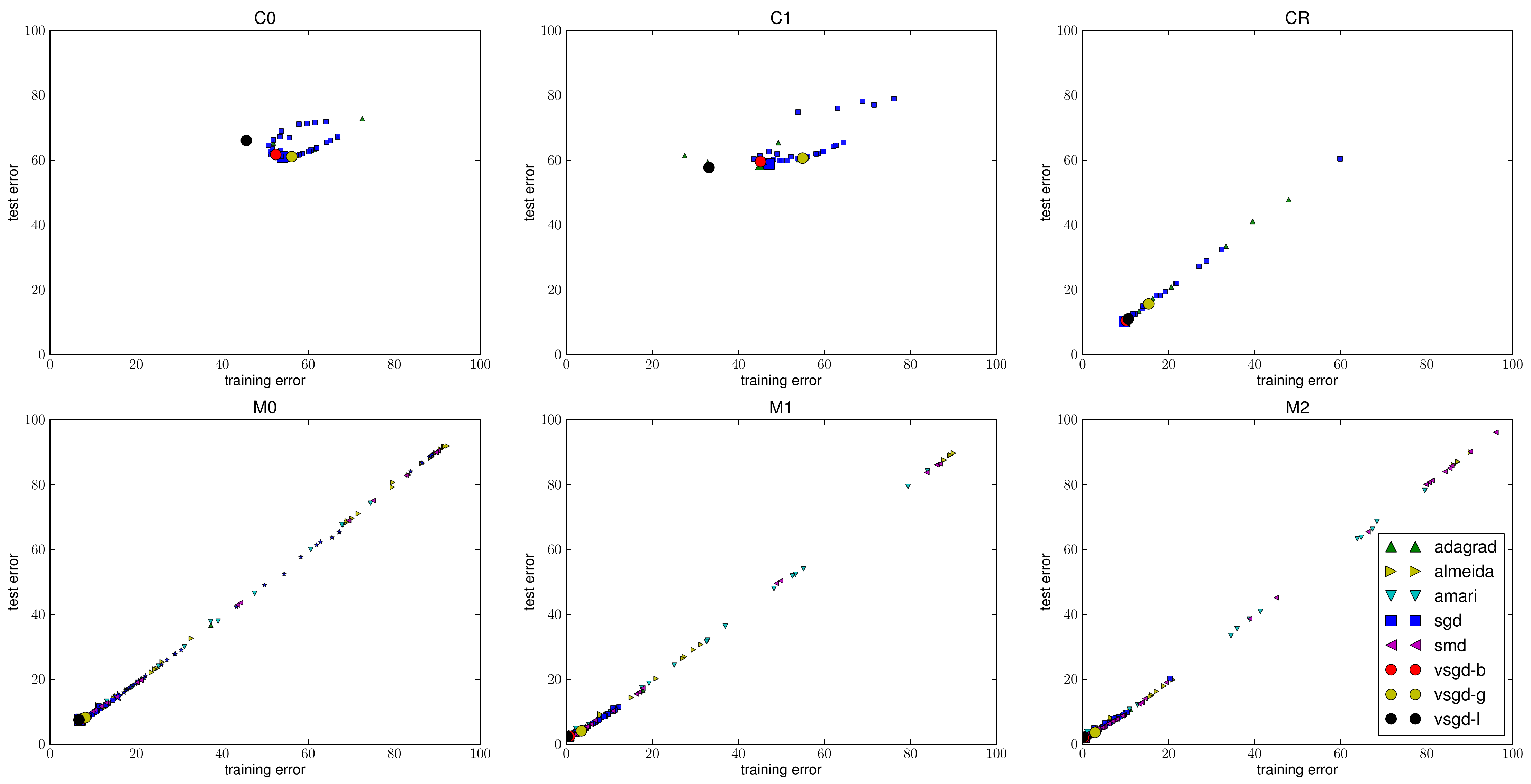}
\caption{Training error versus test error on all 6 setups, global perspective. Different symbol-color combinations correspond to different algorithms, with the best-tuned parameter setting shown as a much larger symbol than the other settings tried.
}
	\label{fig:tvt-glob}
\end{figure*}

\subsection{Non-stationary Quadratic}
In realistic on-line learning scenarios, the curvature or noise level
in any given dimension changes over time (for example because of the
effects of updating other parameters), and thus the learning rates
need to \emph{increase} as well as increase.  Of course, no fixed learning
rate or fixed cooling schedule can achieve this.
To illustrate this, we use again a noisy quadratic loss function, but with abrupt
changes of the optimum every 300 timesteps.

\setlength{\tabcolsep}{4pt}
	
\begin{table*}[htb]
	\centering
	\begin{tabular}{c|cc|cc|c|cc|ccc|cc}
		 & Loss  & Network layer & 
		 \multicolumn{2}{|c|}{SGD}  & \adagrad & \multicolumn{2}{|c|}{Amari}
		 & \multicolumn{3}{|c|}{SMD} &\multicolumn{2}{c}{Almeida}
		  \\
		 &  & sizes &  $\eta_0$ & $\gamma$  & $\eta_0$& $\eta_0$ & $\tau$ 
		  & $\eta_0$ & $\mu$  & $\tau$ 
		  & $\eta_0$& $\tau$
		  \\
		\hline
		\hline
		M0 & CE  & [784, 10]  
		& $3\cdot 10^{-2}$ & $1$      &  $10^{-1}$    
		& $10^{-5}$ & $10^{4}$
		& $10^{-1}$ & $10^{-3}$ & $10^{3}$
		& $10^{-3}$ & $10^{3}$
		\\  
		M1 &  & [784, 120, 10]	           
		& $3\cdot 10^{-2}$  & $1/2 $   &  $10^{-1}$      
		& $10^{-6}$ & $ 5\cdot10^{3}$
		& $3\cdot 10^{-2}$ & $10^{-4}$ & $10^{4}$
		& $10^{-3}$ & $10^{4}$
		\\ 
		M2 &  & [784, 500, 300, 10]	    
		& $10^{-2} $& $1/2$ &  $3\cdot 10^{-2}$ 
		& $3\cdot 10^{-7}$ & $ 5\cdot10^{3}$
		& $3\cdot 10^{-2}$ & $10^{-3}$ & $10^{2}$
		& $10^{-3}$ & $10^{4}$
		\\ 
		\hline
		C0 & CE   & [3072, 10] 
		& $3\cdot 10^{-3}$ & $1$ &  $10^{-2}$ 
		 \\ 
		C1 &  & [3072, 360, 10] 
		& $10^{-2} $& $1$ &  $3\cdot 10^{-3}$  
		\\ 
		CR  & MSE & [3072, 120, 3072] 
		& $3\cdot 10^{-3}$ & $1$ &  $10^{-2}$ 
		\\ 
	\end{tabular}
	\caption{Experimental setup for standard datasets MNIST and
          and the subset of CIFAR-10 using neural
          nets with 0 hidden layer (M0 and C0), 1 hidden layer (M1, C1 and CR), 2 hidden layers (M2). 
          Columns 4 through 13 give the best found hyper-parameters for SGD and the four adaptive algorithms used to compare vSGD to. Note that those hyper-parameters vary substantially across the benchmark tasks.
         }
	\label{tab:setup-prac}
\end{table*}

Figure~\ref{fig:nonstation} shows how vSGD with its adaptive
memory-size appropriately handles such cases. Its initially large
learning rate allows it to quickly approach the optimum, then it
gradually reduces the learning rate as the gradient variance increases
relative to the squared norm of the average gradient, thus allowing the
parameters to closely approach the optimum. When the data
distribution changes (abruptly, in our case), the algorithm automatically detects that
the norm of the average gradient increased relative to the
variance. The learning rate jumps back up and adapts to the new
circumstances.
Note that here and in section~\ref{sec:quad} the curvature is always 1, 
which implies that the preconditioning by the diagonal Hessian component vanishes, 
and still the advantage of adaptive learning rates is clear.

\setlength{\tabcolsep}{8.5pt}

\begin{table*}[htb]
	\centering
\begin{tabular}{l|ccc|c|cccc}
  &vSGD-l &vSGD-b &vSGD-g &SGD &\adagrad &SMD &Amari &Almeida\\ 
\hline
M0 & {\bf 6.72}\% & $7.63$\% & $8.20$\% & $7.05$\% & $6.97$\% & $7.02$\% & $7.33$\% & $11.80$\%\\ 
M1 & {\bf 0.18}\% & $0.78$\% & $3.50$\% & $0.30$\% & $0.58$\% & $0.40$\% & $2.91$\% & $8.49$\%\\ 
M2 & {\bf 0.05}\% & $0.33$\% & $2.91$\% & $0.46$\% & $0.41$\% & $0.55$\% & $1.68$\% & $7.16$\%\\ 
\hline 
C0 & {\bf 45.61}\% & $52.45$\% & $56.16$\% & $54.78$\% & $54.36$\% & --  & --  & -- \\ 
C1 & {\bf 33.16}\% & $45.14$\% & $54.91$\% & $47.12$\% & $45.20$\% & --  & --  & -- \\ 
\hline 
CR & $10.64$ & $10.13$ & $15.37$ & {\bf 9.77} & {\bf 9.80} & --  & --  & -- \\ 
\end{tabular}
	\caption{Final classification error (and reconstruction error
          for CIFAR-2R) on the {\bf training} set, obtained after 6 epochs
          of training, and averaged over ten random initializations.
	Variants are marked in bold if they don't differ statistically significantly from the best one ($p=0.01$).
	Note that the tuning parameters of SGD, \adagrad, SMD, Amari and Almeida are different for each benchmark (see Table~\ref{tab:setup-prac}). 
	We observe the best results with the full element-wise learning rate adaptation (`vSGD-l'), almost always significantly better than the best-tuned SGD or best-tuned \adagrad.
          }
	\label{tab-train}
\end{table*}

\begin{table*}[htb]
	\centering	
	\begin{tabular}{l|ccc|c|cccc}
  &vSGD-l &vSGD-b &vSGD-g &SGD &\adagrad &SMD &Amari &Almeida\\ 
\hline
M0 & {\bf 7.50}\% & $7.89$\% & $8.20$\% & {\bf 7.60}\% & {\bf 7.52}\% & {\bf 7.57}\% & {\bf 7.69}\% & $11.13$\%\\ 
M1 & {\bf 2.42}\% & {\bf 2.44}\% & $4.14$\% & {\bf 2.34}\% & $2.70$\% & {\bf 2.37}\% & $3.95$\% & $8.39$\%\\ 
M2 & {\bf 2.16}\% & {\bf 2.05}\% & $3.65$\% & {\bf 2.15}\% & $2.34$\% & {\bf 2.18}\% & $2.97$\% & $7.32$\%\\ 
\hline 
C0 & $66.05$\% & $61.70$\% & {\bf 61.10}\% & {\bf 61.06}\% & {\bf 61.25}\% & --  & --  & -- \\ 
C1 & {\bf 57.72}\% & $59.55$\% & $60.62$\% & $58.85$\% & $58.67$\% & --  & --  & -- \\ 
\hline 
CR & $11.05$ & $10.57$ & $15.71$ & {\bf 10.29} & {\bf 10.33} & --  & --  & -- \\ 
\hline 
\hline 
 \#settings& 1 & 1 & 1 & 68 & 17 & 476 & 119 & 119 \end{tabular}
	\caption{Final classification error (and reconstruction error
          for CIFAR-2R) on the {\bf test} set, after 6 epochs of training,
          averaged over ten random initializations.
	Variants are marked in bold if they don't differ statistically significantly from the best one ($p=0.01$).
	Note that the parameters of SGD, \adagrad, SMD, Amari and Almeida were finely tuned, on this same test set, and were found to be different for each benchmark (see Table~\ref{tab:setup-prac}); 
	the last line gives the total number of parameter settings over which the tuning was performed.
	Compared to training error, test set performance is more balanced, with vSGD-l being better or statistically equivalent to the best-tuned SGD in 4 out of 6 cases. The main outlier (C0) is a case where the more aggressive element-wise learning rates led to overfitting (compare training error in Table~\ref{tab-train}).
	}
	\label{tab-test}
\end{table*}

\subsection{Neural Network Training}
\label{sec-benchmarks}
SGD is one of the most common training algorithms in use for (large-scale) neural network training.
The experiments in this section compare the three vSGD variants introduced above
with SGD, and some adaptive algorithms described in section~\ref{sec:background} (\adagrad, Almeida, Amari and SMD), with additional details in the appendix.

We exhaustively search for the best hyper-parameter settings among
$\eta_0 \in \{10^{-7}, 3\cdot10^{-7}, 10^{-6}, \ldots, 3\cdot10^0, 10^1\}$, 
$\gamma \in \{0, 1/3, 1/2, 1 \} / \mbox{\#traindata}$,
$\tau \in \{10^{5}, 5\cdot10^{4},10^{4}, 5\cdot10^{3}, 10^{3}, 10^{2},10^{1},  \}$ and
$\mu \in \{10^{-4},10^{-3}, 10^{-2}, 10^{-1}\}$
as determined by their lowest test error (averaged over 2 runs), for each individual benchmark setup.
The last line in Table~\ref{tab-test} shows the total number of settings over which the tuning is done.

\subsubsection{Datasets}
We choose two widely used standard datasets 
to test the different algorithms; the MNIST digit recognition dataset~\cite{lecun-cortes-98} (with 60k training
samples, and 10k test samples), and the CIFAR-10 small natural image dataset~\cite{cifar10}, 
namely the `batch1' subset, which contains 10k training samples and 10k test samples.
We use CIFAR both to learn image classification and reconstruction. 
The only form of preprocessing used (on both datasets) is to normalize the data by substracting mean of the training data along each input dimension.

\subsubsection{Network Architectures}
We use four different architectures of feed-forward neural networks.
\begin{itemize}
\item 
The first one is simple softmax regression (i.e., a network with no hidden layer) for multi-class classification. It has \emph{convex} loss (cross-entropy) relative to parameters. This setup is denoted `M0' for the MNIST case, and `C0' for the CIFAR classification case.
\item 
The second one (denoted `M1'/`C1') is a fully connected multi-layer perceptron, with a single hidden layers, with tanh non-linearities at the hidden units. The cross-entropy loss function is non-convex. 

\item 
The third  (denoted `M2', only used on MNIST) is a deep, fully connected multi-layer perceptron, with a two hidden layers, again with tanh non-linearities. 

\item 
The fourth architecture is a simple autoencoder (denoted `CR'), with one hidden layer (tanh non-linearity) and non-coupled reconstruction weights.
This is trained to minimize the mean squared reconstruction error. Again, the loss is non-convex w.r.t.~the parameters.
\end{itemize}

\begin{figure}[tb]
\centering
\includegraphics[width=0.92\linewidth]{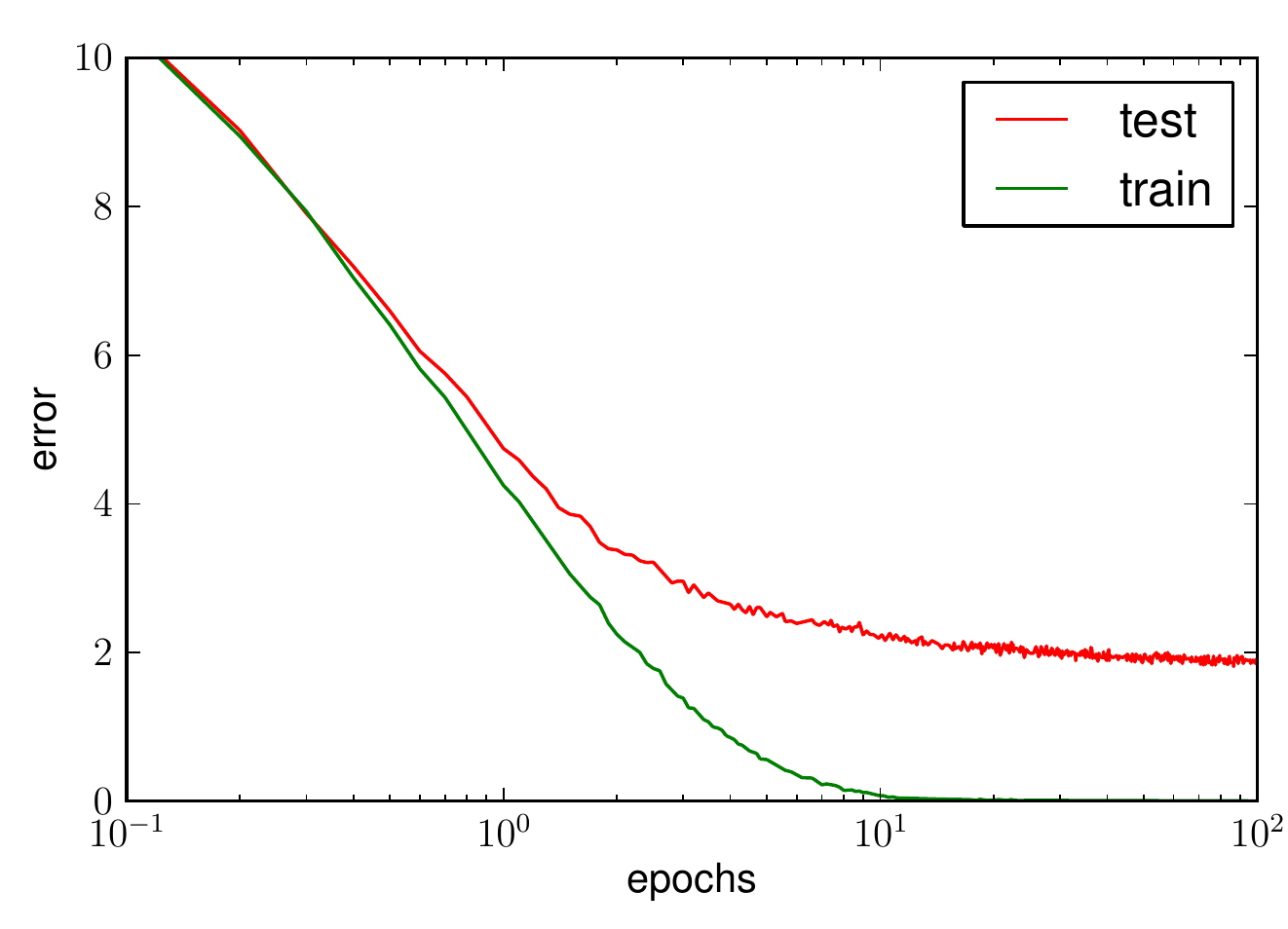}
\caption{Learning curves for full-length runs of 100 episodes, using vSGD-l on the M1 benchmark with 800 hidden units. Test error is shown in red, training error is green. Note the logarithmic scale of the horizontal axis. The average test error after 100 epochs is 1.87\%.
}
	\label{fig:100ep}
\end{figure}

Formally, given input data $h_0=x$, the network processes sequentially through $H>0$ hidden layers by applying affine transform then an element-wise tanh, 
\begin{equation*}
	h_{k+1} = \mbox{tanh} (W_{k} h_{k} + b_{k}),  \quad 
	k = 0, \cdots, H-1.
\end{equation*}
The output of the network $y=h_{H+1} = W_H h_H + b_H$ is then feed into the loss function. 
For cross-entropy loss, the true label $c$ gives the target (delta) distribution to approximate, thus the loss is 
\begin{equation*}
	 \mathbb{E} [ KL ( \delta_c || p_y  ) ]  =  \mathbb{E} [- \log (p_y(c)) ], 
\end{equation*}
where
\begin{equation*}
	p_y(c) = \frac{\exp^{-y(c)}}{\sum_k \exp^{-y(k)}} .
\end{equation*}
For mean-squared reconstruction error, the loss is
\begin{equation}
	\mathbb{E} [  \frac{1}{2}  || x - y ||_2^2  ] 
\end{equation}
The exact numbers of hidden units in each layer, and the corresponding total problem dimensions 
are given in Table~\ref{tab:setup-prac}. The parameters are initialized randomly based on~\citet{glorot-aistats-10}.

To avoid over-fitting, especially for CIFAR which has a comparatively small dataset, we add $\frac{\lambda}{2} ||w||_2^2$, 
a $L^2$ regularization term  on the weights,  to the loss in all experiments (with $\lambda = 10^{-4}$). 
This also avoids numerical instability in vSGD-l, because the estimated diagonal Hessian elements will almost never be close to zero.

\begin{figure*}[htp]
\centering
\includegraphics[width=0.9\linewidth]{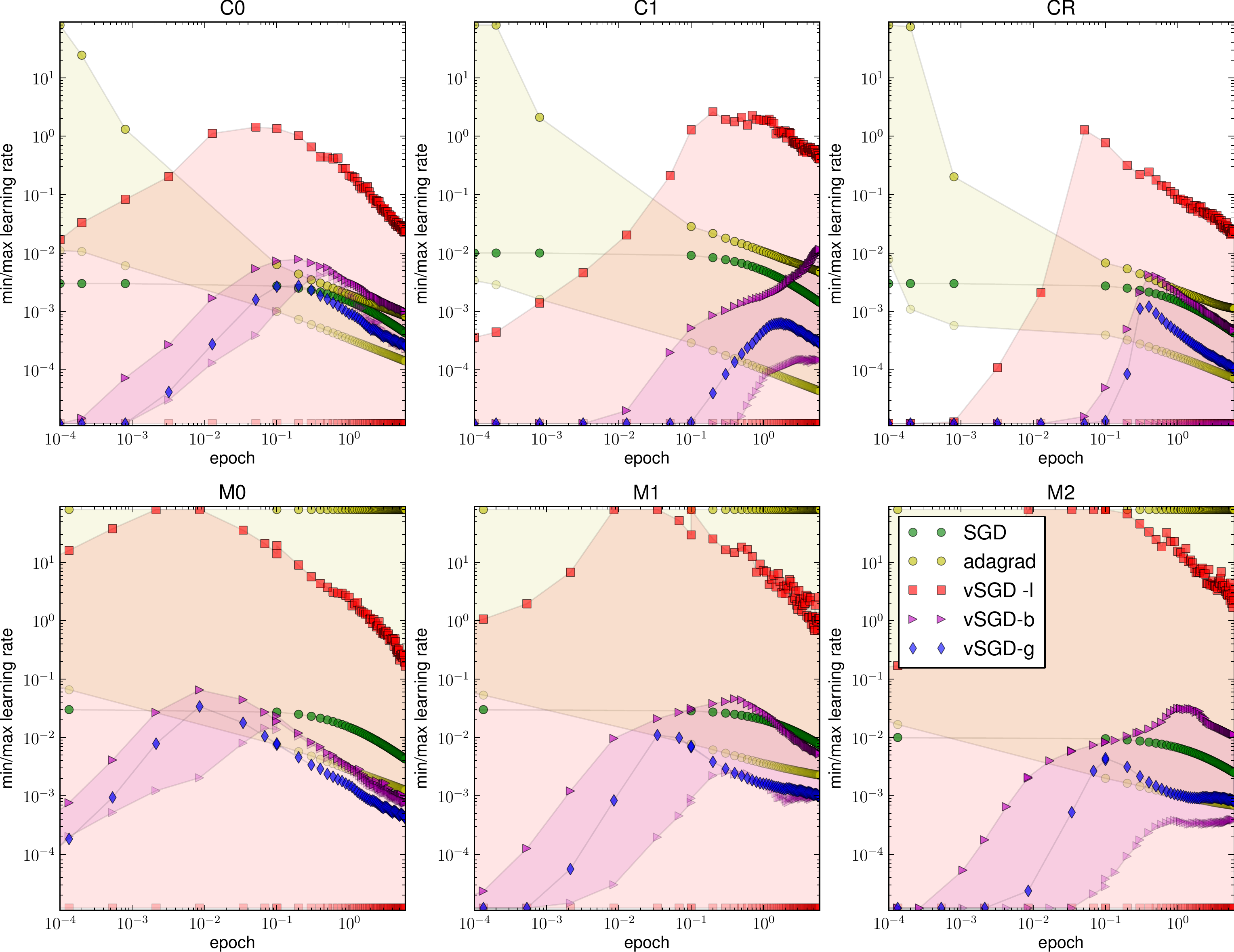}
\caption{Evolution of learning rates. It shows how the learning rates (minimum and maximum across all dimensions) vary as a function of the epoch. Left: CIFAR classification (no hidden layer), right: MNIST classification (no hidden layer).
Each symbol/color corresponds to the median behavior of one algorithm. The range of learning rates (for those algorithms that don't have a single global learning rate) is shown in a colored band in-between the min/max markers. 
The log-log plot highlights the initial behavior, namely the `slow start' (until about 0.1 epochs) due to a large $C$ constant in out methods, which contrasts with the quick start of \adagrad. We also note that \adagrad (yellow circles) has drastically different ranges of learning rates on the two benchmarks.
}
	\label{fig:lr-evo}
\end{figure*}

\subsubsection{Results}
For each benchmark, ten independent runs are averaged and reported in Table~\ref{tab-train} (training set) and Table~\ref{tab-test} (test set). 
They show that the best vSGD variant, across the board, is vSGD-l, which most aggressively adapts one learning rate per dimension. It is almost always significantly better than the best-tuned SGD or best-tuned \adagrad\ in the training set, and
 better or statistically equivalent to the best-tuned SGD in 4 out of 6 cases on the test set. 
 The main outlier (C0) is a case where the more aggressive element-wise learning rates led to overfitting (compare training error in Table~\ref{tab-train}), probably because of the comparatively small dataset size.
Figure~\ref{fig:tvt} illustrates the sensitivity to hyper-parameters of SGD, \adagrad, SMD and Amari's natural gradient on the three MNIST benchmarks: different settings scatter across the performance scale adn tuning matters. This is in stark contrast with vSGD, which without tuning 
obtains the same performance than the best-tuned algorithms.
Figure~\ref{fig:tvt-cifar} does the same for the three CIFAR benchmarks, and Figure~\ref{fig:tvt-glob} provides a more global perspective (zoomed out from the region of interest). 

Figure~\ref{fig:lr-evo} shows the evolution of (minimal/maximal) learning rates over time, emphasizing the 
effects of slow-start initialization in our approach, and Figure~\ref{fig:100ep} shows the learning curve over 100 epochs, much longer than the remainder of the experiments.

\section{Conclusions}
Starting from the idealized case of quadratic loss contributions from
each sample, we derived a method to compute an optimal learning rate
at each update, and (possibly) for each parameter, that optimizes the
expected loss after the next update. The method relies on the square
norm of the expectation of the gradient, and the expectation of the
square norm of the gradient.
We showed different ways of approximating those learning
rates in linear time and space in practice. The experimental results
confirm the theoretical prediction: the adaptive learning rate method
completely eliminates the need for manual tuning of the learning
rate, or for systematic search of its best value.

Our adaptive approach makes SGD more robust in two related ways:
(a) When used in on-line training
scenarios with non-stationary signals, the adaptive learning rate
automatically increases when the distribution changes, so as to adjust
the model to the new distribution, and automatically decreases in
stable periods when the system fine-tunes itself within an attractor.
This provides robustness to dynamic changes of the optimization landscape.
(b) The tuning-free property implies that the same algorithm can
adapt to drastically different circumstances, which can appear within a 
single (deep or heterogeneous) network. This robustness alleviates 
the need for careful normalizations of inputs and structural components.

Given the successful validation on a variety of classical
large-scale learning problems, we hope that this enables for SGD to
be a truly user-friendly `out-of-the-box' method.


\subsubsection*{Acknowledgments}
The authors want to thank Camille Couprie, Cl\'{e}ment Farabet and
Arthur Szlam for helpful discussions, and Shane
Legg for the paper title. This work was funded in part through AFR
postdoc grant number 2915104, of the National Research Fund
Luxembourg, and ONR Grant 5-74100-F6510.

\newpage

\appendix
\section{Convergence Proof}
If we do gradient descent with $\eta^*(t)$, then almost surely, the algorithm
converges (for the quadratic model). 
To prove that, we follow
classical techniques based on Lyapunov stability theory
\cite{bucy-65}. Notice that the expected loss follows
\begin{eqnarray*}
&&\Expectation \left[J\left(\theta^{(t+1)}\right) \; | \;\theta^{(t)}\right] \\
&=&  \frac{1}{2}h \cdot
\Expectation \left[\left( (1-\eta^{*} h  )(\theta^{(t)} -\theta^*) + \eta^{*} h \sigma   \xi  \right)^2 
+\sigma^2
 \right]\nonumber\\
&=& \frac{1}{2}h 
\left[
(1-\eta^{*} h  )^2 (\theta^{(t)} -\theta^*)^2  + (\eta^{*})^2 h^2 \sigma^2 +\sigma^2
\right]\\
&=&\frac{1}{2}h 
\left[ \frac{\sigma^2} { (\theta^{(t)} -\theta^*)^2 + \sigma^2} (\theta^{(t)} -\theta^*)^2 +\sigma^2  \right] \\
&\le& J\left(\theta^{(t)}\right) 
\end{eqnarray*}
Thus $J(\theta^{(t)})$ is a positive super-martingale, indicating that
almost surely $ J(\theta^{(t)}) \to J^{\infty}$.
We are to prove that almost surely $J^{\infty} = J(\theta^*) =
\frac{1}{2}h\sigma^2$. Observe that
\begin{eqnarray*}
  J(\theta^{(t)}) - \Expectation [ J(\theta^{(t+1)})  \mid \theta^{(t)} ] & =&  \frac{1}{2}h\eta^*(t) \;,
\\
  \Expectation [  J(\theta^{(t)})] - \Expectation [ J(\theta^{(t+1)})  \mid \theta^{(t)} ]      
   &=&  \frac{1}{2}h \Expectation[\eta^*(t)]
\end{eqnarray*}

Since $ \Expectation [ J(\theta^{(t)}) ] $ is bounded below by 0, the
telescoping sum gives us $\Expectation [\eta^*(t)] \to 0$, which in
turn implies that in probability $\eta^*(t) \to 0$.  We can rewrite
this as
\begin{equation*}
 \eta^*(t) =  \frac{ J(\theta_{t}) -\frac{1}{2}h\sigma^2}{J(\theta_{t}) }  \to 0
\end{equation*}
By uniqueness of the limit, almost surely, $\frac{ J^\infty
  -\frac{1}{2}h\sigma^2}{ J^\infty } = 0$.  Given that $J$ is strictly
positive everywhere, we conclude that $ J^{\infty} =
\frac{1}{2}h\sigma^2$ almost surely, i.e $ J(\theta^{(t)}) \to
\frac{1}{2}h\sigma^2 = J(\theta^*)$.

\begin{figure}[tb]
\centering
\includegraphics[width=0.92\linewidth]{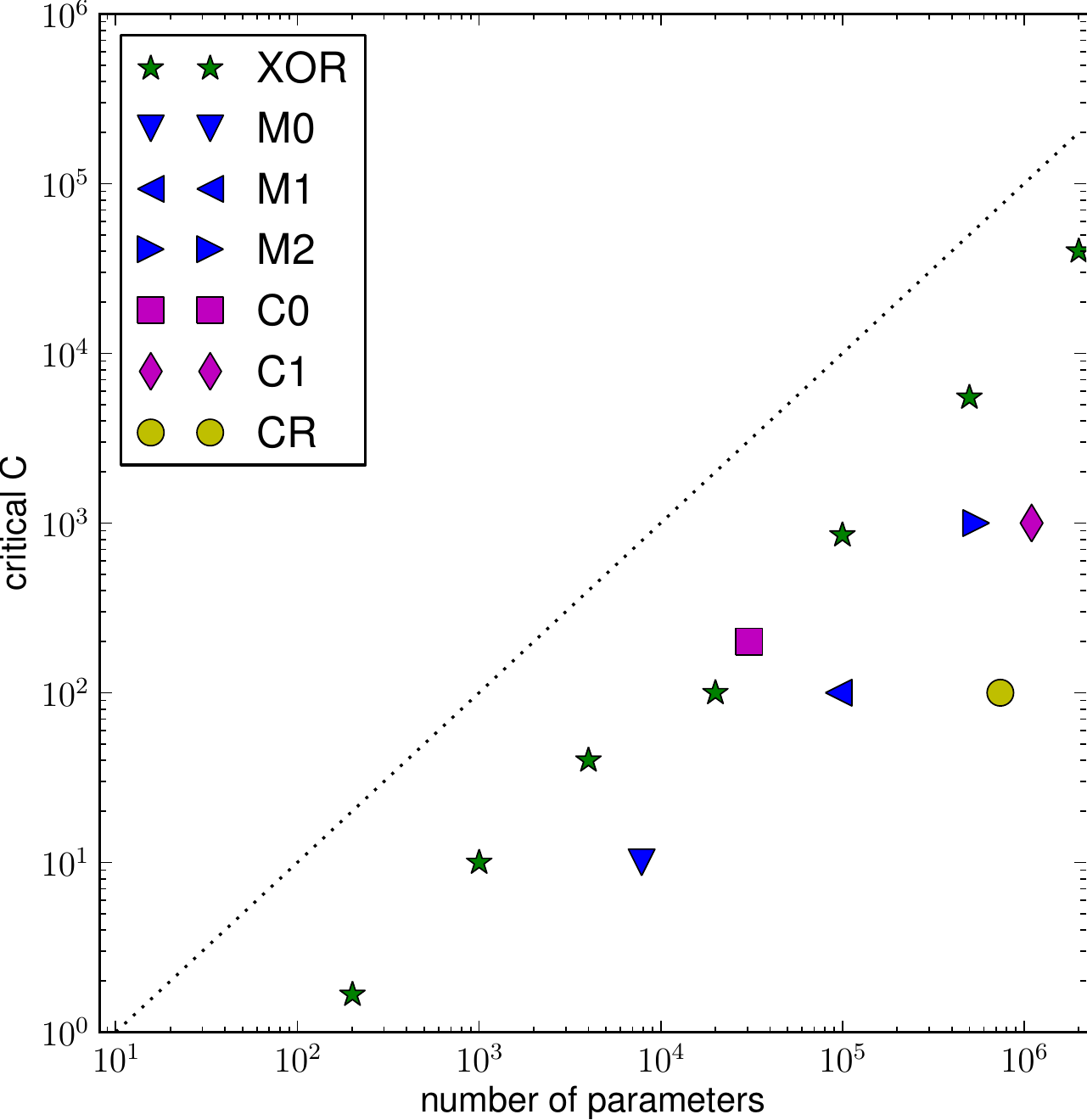}
\caption{Critical values for initialization parameter $C$. 
This plot shows the values of $C$ below which vSGD-l becomes unstable (too large initial steps).
We determine the critical $C$ value as the largest for which at least 10\% of the runs give rise to instability.
The markers correspond to experiments with setups on a broad range of parameter dimensions. 
Six markers correspond to the benchmark setups from the main paper, and the green stars 
correspond to simple the XOR-classification task with an MLP of a single hidden layer, the size of which is varied from 2 to 500000 neurons.
The black dotted diagonal line indicates, our `safe' heuristic choice of $C=d/10$.
}
	\label{fig:c-bound}
\end{figure}

\begin{figure*}[htp]
\centering
\includegraphics[width=\linewidth]{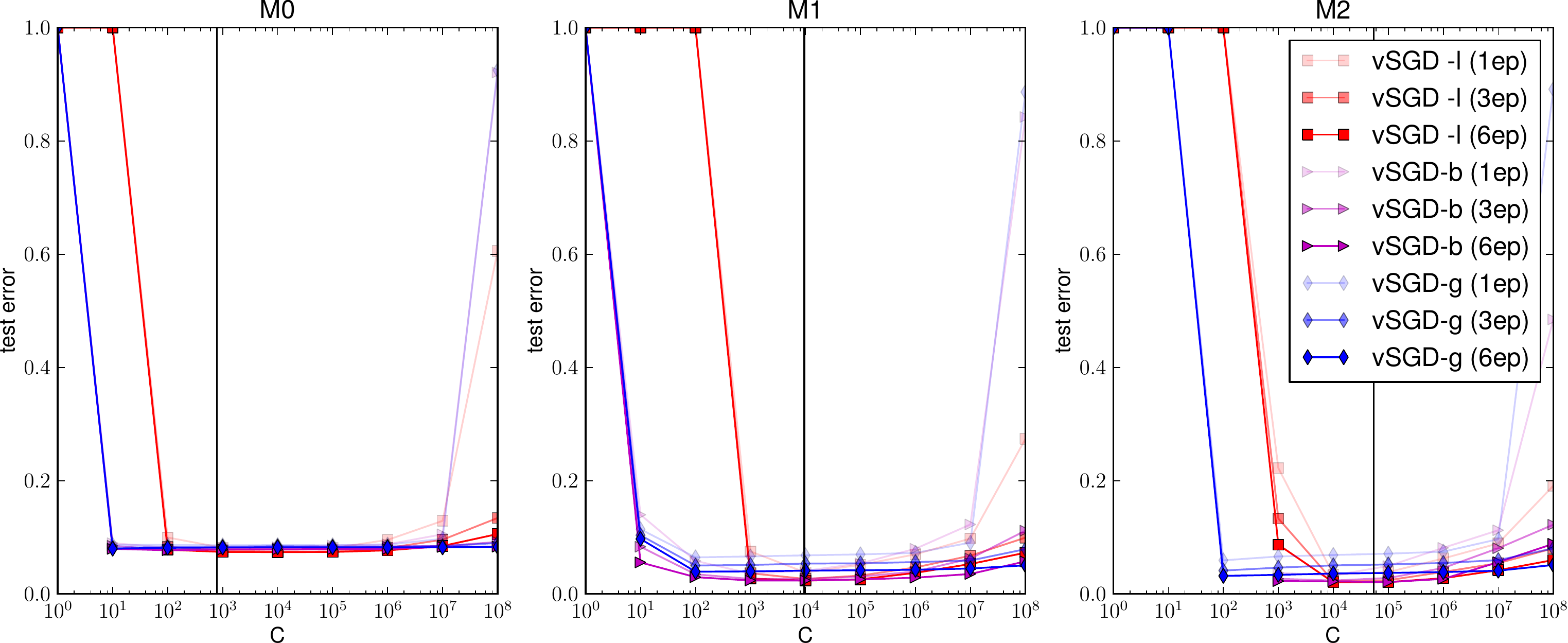}

\caption{Parameter study on hyper-parameter $C$. These plots demonstrate that the algorithm is 
insensitive to the choice of initial slowness parameter $C$. For each of the setups on the MNIST classification benchmark (with vastly differing parameter dimension $d$, see Table 1 in the main paper,
we show the sensitivity of the test set performance as we vary $C$ over 8 orders of magnitude. 
Each plot shows the test errors after 1, 3 and 6 epochs (different levels of transparency), for the three adaptive variants (l, b, g, in different colors).
In all cases, we find that the 
updates are unstable if $C$ is chosen too small (the element-wise `l' variant being most affected), but otherwise $C$ has very little effect, up until when it becomes extremely large: for $C=10^8$, this initialization basically neutralizes the whole first epoch, and is still felt at epoch 6.
The black vertical line indicates, for the three setups, 
our `safe' heuristic choice of $C=d/10$. Its only purpose is to avoid instability upon initialization, and given that its 'sweet spot' spans many orders of magnitude, it does not need to be tuned more precisely.
}
	\label{fig:c-study}
\end{figure*}

\section{Derivation of Global Learning Rate}
We can derive an optimal global learning rate
$\eta^*_g$ as follows.
\begin{eqnarray*}
\eta_g^* (t)
&=& 
\arg\min_{ \eta}
\Expectation \left[\mathcal{J}\left(\params^{(t+1)}\right) \; | \;\params^{(t)}\right]\\
&=&
\arg\min_{\eta}
\sum_{i=1}^d h_i \left(
(1-\eta h_i  )^2 (\theta_i^{(t)} -\theta^*_i)^2 
\right. \\&& \hspace{3.5cm}\left.
+\sigma_i^2 +\eta^2 h_i^2 \sigma_i^2
\right)
\\&=&
\arg\min_{\eta}
\left[\eta^2 \sum_{i=1}^d \left(
h_i^3 (\theta_i^{(t)} -\theta^*_i)^2 
+ h_i^3 \sigma_i^2\right)
\right. \\&& \hspace{3cm}\left.
-2\eta  \sum_{i=1}^d h^2_i  (\theta_i^{(t)} -\theta^*_i)^2 \right]
\end{eqnarray*}
which gives
\begin{eqnarray*}
\eta_g^*(t)
&=& 
\frac{\sum_{i=1}^d h^2_i  (\theta_i^{(t)} -\theta^*_i)^2}
{ \sum_{i=1}^d \left(
h_i^3 (\theta_i^{(t)} -\theta^*_i)^2 
+  h_i^3 \sigma_i^2\right)
}
\end{eqnarray*}

The adaptive time-constant for the global case is:
\begin{eqnarray*}
\tau_g (t+1) &=& \left(1-\frac{\sum_{i=1}^{d}\hatg^2}{\hatvg(t)}\right) \cdot  \tau_g(t) \;+\; 1 
\label{eq:memory-update-glob}
\end{eqnarray*}

\section{SMD Implementation}
The details of our implementation of SMD (based on a global learning rates) are given by the following updates:
\begin{eqnarray*}
	\params_{t+1} &\leftarrow& \params_{t} - \eta_t \grad \\
	\eta_{t+1} &\leftarrow& \eta_t \exp\left(-\mu\grad\transp\mathbf{v}_{t}\right) \\
	\mathbf{v}_{t+1} &\leftarrow & (1- \tau^{-1}) \mathbf{v}_{t} - \eta_{t} \left(\grad + (1-\tau^{-1}) \cdot \mathbf{H}_{t}\mathbf{v}_{t}\right)
\end{eqnarray*}
where $\mathbf{H}\mathbf{v}$ denotes the Hessian-vector product with vector $\mathbf{v}$, which can be computed in linear time.
The three hyper-parameters used are the initial learning rate $\eta_0$, the meta-learning rate $\mu$, and a time constant $\tau$ for updating the auxiliary vector $\mathbf{v}$.

\section{Sensitivity to Initialization}
Figure~\ref{fig:c-study} shows that the  initialization parameter $C$ does not affect performance, so long as it is sufficiently large.
This is not surprising, because its only effect is to slow down the initial step sizes until accurate 
exponential averages of the interesting quantities can be computed. 

There is a \emph{critical} minimum value of $C$, blow which the algorithm is unstable.
Figure~\ref{fig:c-bound} shows what those critical values are for 13 different setups with widely varying problem dimension. 
From these empirical results, we derive our rule-of-thumb choice of $C=d/10$ as a `safe' pick for the constant (in fact it is even a factor 10 larger than the observed critical value for any of the benchmarks, just to be extra careful).

\bibliographystyle{icml2013}
\bibliography{sgd.bib}

\end{document}